\pdfoutput=1

\documentclass[11pt]{article}

\usepackage[]{EMNLP2023}

\usepackage{times}
\usepackage{latexsym}

\usepackage[T1]{fontenc}

\usepackage[utf8]{inputenc}

\usepackage{microtype}

\usepackage{inconsolata}

\usepackage{graphicx}
\usepackage{subcaption}
\usepackage{dblfloatfix} 

\usepackage{enumitem}

\usepackage{algorithm}
\usepackage{listings}

\definecolor{codegray}{rgb}{0.5,0.5,0.5}
\definecolor{codepurple}{rgb}{0.58,0,0.82}
\definecolor{codegreen}{rgb}{0,0.58,0.22}

\lstdefinestyle{mystyle}{
    commentstyle=\color{codegray},
    keywordstyle = \color{blue},
    numberstyle=\tiny\color{codegray},
    stringstyle=\color{codegreen},
    basicstyle=\ttfamily\footnotesize,
    breakatwhitespace=false,         
    breaklines=true,                 
    captionpos=b,                    
    keepspaces=true,                 
    numbers=left,                    
    numbersep=5pt,                  
    showspaces=false,                
    showstringspaces=false,
    showtabs=false,                  
    tabsize=1
}

\usepackage{booktabs}

\usepackage{amsmath}

\title{\texttt{signwriting-evaluation}: \\Effective Sign Language Evaluation via SignWriting}

\author{Amit Moryossef$^{1,2}$ \;\; Rotem Zilberman$^3$ \;\; Ohad Langer$^3$ \\
\texttt{amit@sign.mt, \{rotemz.mail,ohadlanger1\}@gmail.com} \\
University of Zurich$^1$, \url{sign.mt}$^2$, Bar-Ilan University$^3$}

\begin{document}
\maketitle

\begin{abstract}
The lack of automatic evaluation metrics tailored for SignWriting presents a significant obstacle in developing effective transcription and translation models for signed languages. This paper introduces a comprehensive suite of evaluation metrics specifically designed for SignWriting, including adaptations of standard metrics such as \texttt{BLEU} and \texttt{chrF}, the application of \texttt{CLIPScore} to SignWriting images, and a novel symbol distance metric unique to our approach. We address the distinct challenges of evaluating single signs versus continuous signing and provide qualitative demonstrations of metric efficacy through score distribution analyses and nearest-neighbor searches within the SignBank corpus. Our findings reveal the strengths and limitations of each metric, offering valuable insights for future advancements using SignWriting. This work contributes essential tools for evaluating SignWriting models, facilitating progress in the field of sign language processing.
Our code is available at \url{https://github.com/sign-language-processing/signwriting-evaluation}.
\end{abstract}

\section{Introduction}

Signed languages are the primary means of communication for deaf communities worldwide, possessing rich linguistic structures fundamentally different from spoken languages. SignWriting is a notation system developed to visually represent signed languages, capturing hand shapes, movements, facial expressions, and body positions \cite{writing:sutton1990lessons}. Despite its potential to enhance sign language processing, the lack of specialized automatic evaluation metrics for SignWriting hinders the development of effective transcription and translation models \cite{moryossef2023signbankplus}.

Traditional evaluation metrics such as \texttt{BLEU} \cite{papineni-etal-2002-bleu}, \texttt{chrF} \cite{popovic-2015-chrf}, and \texttt{CLIPScore} \cite{hessel-etal-2021-clipscore} have been extensively used in natural language processing tasks. 
However, these metrics are well-documented in their struggle to accurately assess the performance of models involving complex relations and meaningful nuances and therefore, they are not directly applicable to the visual-spatial SignWriting \cite{reiter-2018-structured, kim-etal-2024-signbleu, hessel-etal-2021-clipscore}. 
This discrepancy leads to inadequate assessments of model performance and often forces researchers to develop ad-hoc evaluation metrics without proper validation \cite{jiang2022machine}.

To overcome these limitations, we present \texttt{signwriting-evaluation}, a comprehensive toolkit providing a suite of automatic evaluation metrics specifically tailored for SignWriting. Our toolkit adapts standard metrics such as \texttt{BLEU} and \texttt{chrF} for tokenized and untokenized Formal SignWriting (FSW) strings, respectively, enabling their application to sign language data. We also leverage \texttt{CLIPScore} by applying it to SignWriting images using the original CLIP model, bridging the gap between visual representations and textual evaluation. Moreover, we develop a novel symbol distance metric that accounts for the visual and spatial properties of SignWriting symbols, offering a better tailored assessment of transcription and translation models.

We demonstrate the efficacy of our evaluation metrics through qualitative analyses, including score distribution examinations and nearest-neighbor searches within the SignBank corpus, which contains around 230,000 unique single signs. Our findings highlight the strengths and limitations of each metric, providing valuable insights for future advancements in SignWriting processing.

By releasing \texttt{signwriting-evaluation} as an open-source toolkit, we aim to facilitate further research and development in the field of sign language processing. We believe that our work will help create more effective and accurate models for SignWriting transcription and translation.

\section{Evaluation Metrics}

To effectively evaluate SignWriting transcriptions and translations, we explore both traditional metrics (\texttt{BLEU} (\S\ref{sec:metrics:bleu}), \texttt{chrF} (\S\ref{sec:metrics:chrf}), and \texttt{CLIPScore} (\S\ref{sec:metrics:clip})) adapted for SignWriting and introduce a new metric specifically designed to capture its unique characteristics (\S\ref{sec:metrics:symbol_distance}).

\subsection{\texttt{BLEU} \cite{papineni-etal-2002-bleu}}
\label{sec:metrics:bleu}

\texttt{BLEU} (Bilingual Evaluation Understudy) is a common metric for evaluating machine translation by comparing n-grams of the candidate translation with reference translations \cite{blagec2022global}. 
It measures the overlap of n-grams between the hypothesis and reference texts, providing a quantitative measure of their similarity.

To adapt \texttt{BLEU} for SignWriting, as a baseline metric, we tokenize the Formal SignWriting (FSW) strings into sequences of symbols. Each symbol in FSW represents a specific handshape, movement, facial expression, or other sign components, encoded as a combination of characters, which we treat as tokenized sequences. We use the \texttt{BLEU} implementation from \texttt{sacreBLEU} \cite{post-2018-call}.

However, applying \texttt{BLEU} directly to SignWriting presents challenges like requiring exact symbol position matches, where minor shifts cause mismatches and lower scores despite minimal visual differences \cite{reiter-2018-structured}. It also ignores symbol similarity, treating tokens as discrete units. Moreover, the flexible symbol ordering in FSW can lead to lower scores even for the same sign.

\subsection{\texttt{chrF} \cite{popovic-2015-chrf}}
\label{sec:metrics:chrf}

\texttt{chrF} is a character n-gram F-score metric originally designed for machine translation evaluation. It computes the F-score based on the precision and recall of character n-grams between the hypothesis and reference texts. Since Formal SignWriting strings are composed of sequences of characters representing symbols, \texttt{chrF} can be directly applied without tokenization.  We use the \texttt{chrF} implementation from \texttt{sacreBLEU} \cite{post-2018-call}.

Using \texttt{chrF} to capture character-level similarities in SignWriting, can help assess minor symbol variations. However, like \texttt{BLEU}, it overlooks semantic meaning and visual representation of the symbols, potentially limiting its ability to correctly evaluate SignWriting outputs \cite{kim-etal-2024-signbleu}.

\subsection{\texttt{CLIPScore} \cite{hessel-etal-2021-clipscore}}
\label{sec:metrics:clip}

\texttt{CLIPScore} is an evaluation metric using the CLIP (Contrastive Language–Image Pre-training; \citet{radford2021learning}) model to compute the similarity between images and text. We adapt it for SignWriting by visualizing the script and computing the cosine similarity between image embeddings.

We use the original CLIP model without additional training, relying on its ability to capture visual similarities. 
Trained on a large image-text dataset, CLIP identifies visual similarities even when symbols differ, capturing aspects that token-based metrics may overlook.
However, the train/test domain mismatch limits its ability to capture nuances, yielding similar embeddings for all SignWriting images with minor differentiation.

\subsection{Symbol Distance Metric}
\label{sec:metrics:symbol_distance}

To address the limitations of existing metrics above, we propose a novel metric specifically designed for SignWriting notation. This metric accounts for the visual and spatial properties of symbols, providing a more accurate evaluation of transcription and translation outputs.

Our metric assesses the similarity between two SignWriting signs by calculating distances between their constituent symbols based on attributes such as base shape, orientation, rotation, and position. We define a distance function that considers these attributes, weighted by their significance.

To compare two signs as sets of symbols, we use the Hungarian algorithm \cite{Kuhn1955Hungarian} to find the optimal matching that minimizes the total distance between symbols from the hypothesis and reference. 
The \textcolor{ForestGreen}{mean normalized distance $\bar{D}$} is calculated over the matched symbol pairs. We normalize the distances using a non-linear function to ensure they are in a consistent range, exponentially scaled by \textcolor{red}{$\alpha$} which controls the normalization curve.

To account for differences in the number of symbols between signs, we compute a \textcolor{purple}{length adjustment factor}, using the \textcolor{orange}{number of symbols within each sign} and a \textcolor{OrangeRed}{penalty severity control $\beta$}:

\begin{equation}
\textcolor{purple}{L} = \left|\frac{\textcolor{orange}{|S_{\text{hyp}}|} - \textcolor{orange}{|S_{\text{ref}}||}}{\max(\textcolor{orange}{|S_{\text{hyp}}|}, \textcolor{orange}{|S_{\text{ref}}|}) + 1} \right|^{\textcolor{OrangeRed}{\beta }}
\end{equation}

The final score between the hypothesis and reference signs is computed as $(1 - \textcolor{ForestGreen}{\bar{D}})(1 - \textcolor{purple}{L})$, which is then normalized exponentially by parameter \textcolor{brown}{$\gamma$}. This score ranges between $0$ and $1$, with higher values indicating greater similarity.

Our symbol distance metric provides a task-specific evaluation, capturing subtle differences in symbol attributes critical for conveying meaning in sign language. By incorporating both symbol-level distances and structural discrepancies, it offers a balanced and efficient assessment of model outputs.

\section{Qualitative Evaluation}

In this section, we analyze the effectiveness and limitations of the proposed metrics, focusing on both evaluation score distribution (\S\ref{sec:distribution}) and performance in a nearest-neighbor search (\S\ref{sec:nearest_neighbor}).

\subsection{Distribution of Scores}
\label{sec:distribution}

\begin{figure}[ht]
    \centering
    \includegraphics[width=\linewidth]{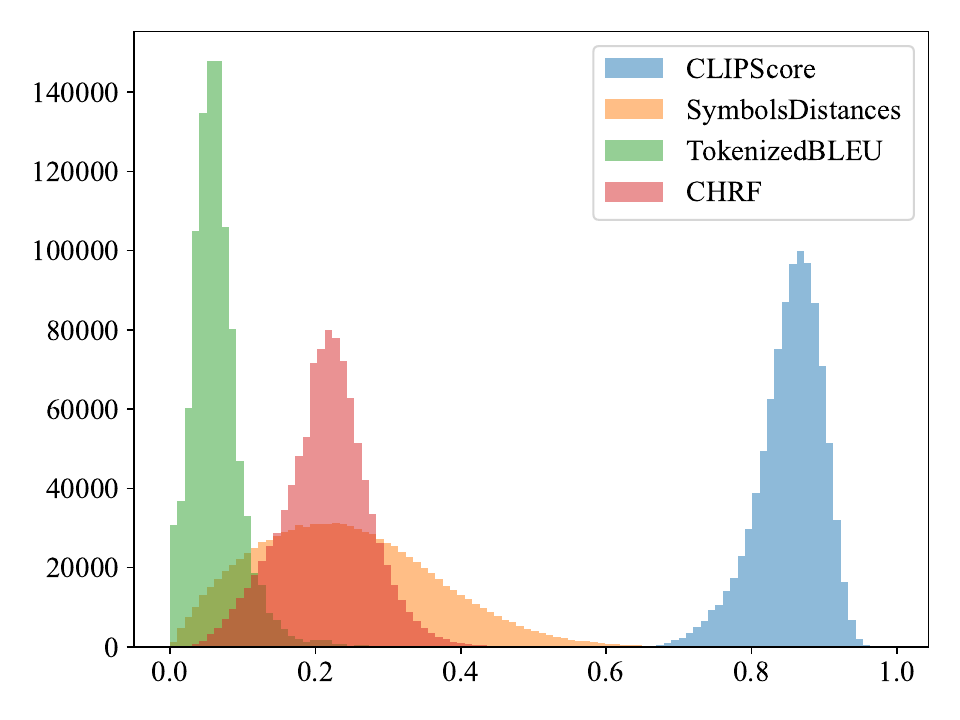}
    \caption{Distribution of scores for random pairs of signs using different evaluation metrics.}
    \label{fig:score_distribution}
\end{figure}

To understand how each metric behaves when comparing random pairs of signs, we computed the any-to-any scores for a random sample of $1,000$ signs from the \emph{SignBank}\footnote{\url{https://www.signbank.org/signpuddle/}} corpus, which includes around $230,000$ unique single signs. Intuitively, we expect random sign pairs to have low similarity scores due to their likely lack of relation. This should be reflected in a score distribution skewed towards lower values. Figure~\ref{fig:score_distribution} shows the histograms of scores for each metric and reveals that:

\begin{itemize}
    \item \textbf{\texttt{BLEU}} scores are concentrated near zero, due to random signs sharing few common tokens.
    \item \textbf{\texttt{chrF}} shows a wider spread, suggesting higher character-level overlap between random signs.
    \item \textbf{\texttt{CLIPScore}} yields scores clustered around a high value, due to the model's tendency to produce similar embeddings for different images \cite{ahmadi-agrawal-2024-examination}.
    \item \textbf{Symbol Distance} results in a wide distribution of low scores, with a tail of higher scores, reflecting the low similarity between most random signs when considering symbol attributes and positions, and higher similarity between fewer signs.
\end{itemize}

These distributions help us understand the sensitivity and discriminative power of each metric in distinguishing between unrelated signs.

\subsection{Nearest Neighbor Search}
\label{sec:nearest_neighbor}

To further assess the practical effectiveness of the metrics, we conducted a nearest-neighbor search on the SignBank corpus. We selected multiple variations for the sign `Hello' and retrievedtop $10$ nearest-neighbors in the entire corpus for each metric.
Table~\ref{tab:nearest_neighbors} shows the ranked results for the three reference signs across four metrics.

\begin{table*}[ht]
    \centering
    \resizebox{\linewidth}{!}{%
    \begin{tabular}{c|cccc|cccc|cccc|}
        & \multicolumn{4}{c|}{\includegraphics[scale=0.8]{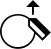}} & \multicolumn{4}{c|}{\includegraphics[scale=0.8]{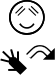}} & \multicolumn{4}{c|}{\includegraphics[scale=0.8]{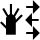}} \\
        \cmidrule{2-13}
        Rank & \texttt{CLIP} & \texttt{Distance} & \texttt{BLEU} & \texttt{chrF} & \texttt{CLIP} & \texttt{Distance} & \texttt{BLEU} & \texttt{chrF} & \texttt{CLIP} & \texttt{Distance} & \texttt{BLEU} & \texttt{chrF} \\
        \midrule
        1 & \includegraphics[scale=0.5]{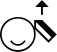} & \includegraphics[scale=0.5]{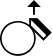} & \includegraphics[scale=0.5]{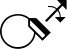} & \includegraphics[scale=0.5]{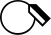} & \includegraphics[scale=0.5]{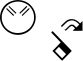} & \includegraphics[scale=0.5]{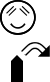} & \includegraphics[scale=0.5]{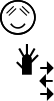} & \includegraphics[scale=0.5]{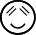} & \includegraphics[scale=0.5]{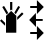} & \includegraphics[scale=0.5]{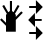} & \includegraphics[scale=0.5]{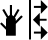} & \includegraphics[scale=0.5]{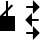} \\
        2 & \includegraphics[scale=0.5]{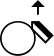} & \includegraphics[scale=0.5]{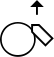} & \includegraphics[scale=0.5]{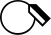} & \includegraphics[scale=0.5]{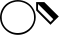} & \includegraphics[scale=0.5]{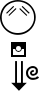} & \includegraphics[scale=0.5]{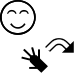} & \includegraphics[scale=0.5]{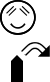} & \includegraphics[scale=0.5]{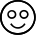} & \includegraphics[scale=0.5]{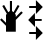} & \includegraphics[scale=0.5]{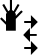} & \includegraphics[scale=0.5]{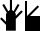} & \includegraphics[scale=0.5]{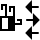} \\
        3 & \includegraphics[scale=0.5]{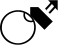} & \includegraphics[scale=0.5]{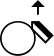} & \includegraphics[scale=0.5]{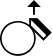} & \includegraphics[scale=0.5]{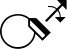} & \includegraphics[scale=0.5]{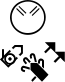} & \includegraphics[scale=0.5]{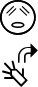} & \includegraphics[scale=0.5]{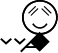} & \includegraphics[scale=0.5]{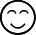} & \includegraphics[scale=0.5]{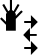} & \includegraphics[scale=0.5]{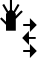} & \includegraphics[scale=0.5]{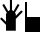} & \includegraphics[scale=0.5]{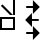} \\
        4 & \includegraphics[scale=0.5]{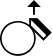} & \includegraphics[scale=0.5]{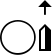} & \includegraphics[scale=0.5]{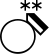} & \includegraphics[scale=0.5]{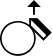} & \includegraphics[scale=0.5]{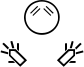} & \includegraphics[scale=0.5]{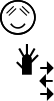} & \includegraphics[scale=0.5]{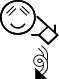} & \includegraphics[scale=0.5]{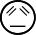} & \includegraphics[scale=0.5]{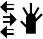} & \includegraphics[scale=0.5]{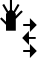} & \includegraphics[scale=0.5]{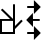} & \includegraphics[scale=0.5]{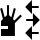} \\
        5 & \includegraphics[scale=0.5]{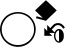} & \includegraphics[scale=0.5]{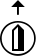} & \includegraphics[scale=0.5]{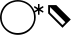} & \includegraphics[scale=0.5]{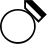} & \includegraphics[scale=0.5]{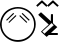} & \includegraphics[scale=0.5]{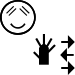} & \includegraphics[scale=0.5]{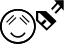} & \includegraphics[scale=0.5]{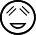} & \includegraphics[scale=0.5]{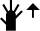} & \includegraphics[scale=0.5]{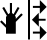} & \includegraphics[scale=0.5]{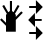} & \includegraphics[scale=0.5]{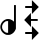} \\
        6 & \includegraphics[scale=0.5]{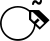} & \includegraphics[scale=0.5]{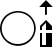} & \includegraphics[scale=0.5]{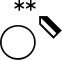} & \includegraphics[scale=0.5]{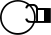} & \includegraphics[scale=0.5]{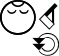} & \includegraphics[scale=0.5]{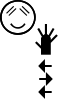} & \includegraphics[scale=0.5]{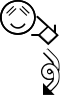} & \includegraphics[scale=0.5]{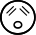} & \includegraphics[scale=0.5]{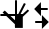} & \includegraphics[scale=0.5]{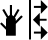} & \includegraphics[scale=0.5]{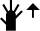} & \includegraphics[scale=0.5]{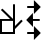} \\
        7 & \includegraphics[scale=0.5]{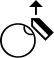} & \includegraphics[scale=0.5]{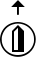} & \includegraphics[scale=0.5]{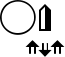} & \includegraphics[scale=0.5]{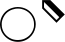} & \includegraphics[scale=0.5]{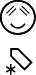} & \includegraphics[scale=0.5]{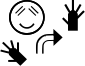} & \includegraphics[scale=0.5]{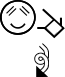} & \includegraphics[scale=0.5]{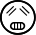} & \includegraphics[scale=0.5]{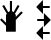} & \includegraphics[scale=0.5]{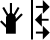} & \includegraphics[scale=0.5]{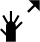} & \includegraphics[scale=0.5]{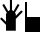} \\
        8 & \includegraphics[scale=0.5]{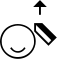} & \includegraphics[scale=0.5]{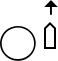} & \includegraphics[scale=0.5]{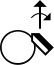} & \includegraphics[scale=0.5]{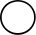} & \includegraphics[scale=0.5]{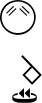} & \includegraphics[scale=0.5]{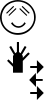} & \includegraphics[scale=0.5]{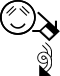} & \includegraphics[scale=0.5]{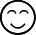} & \includegraphics[scale=0.5]{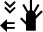} & \includegraphics[scale=0.5]{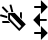} & \includegraphics[scale=0.5]{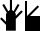} & \includegraphics[scale=0.5]{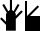} \\
        9 & \includegraphics[scale=0.5]{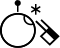} & \includegraphics[scale=0.5]{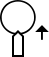} & \includegraphics[scale=0.5]{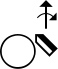} & \includegraphics[scale=0.5]{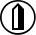} & \includegraphics[scale=0.5]{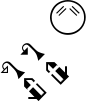} & \includegraphics[scale=0.5]{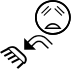} & \includegraphics[scale=0.5]{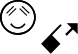} & \includegraphics[scale=0.5]{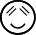} & \includegraphics[scale=0.5]{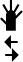} & \includegraphics[scale=0.5]{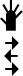} & \includegraphics[scale=0.5]{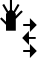} & \includegraphics[scale=0.5]{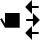} \\
        10 & \includegraphics[scale=0.5]{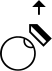} & \includegraphics[scale=0.5]{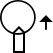} & \includegraphics[scale=0.5]{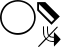} & \includegraphics[scale=0.5]{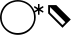} & \includegraphics[scale=0.5]{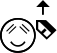} & \includegraphics[scale=0.5]{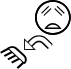} & \includegraphics[scale=0.5]{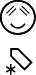} & \includegraphics[scale=0.5]{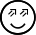} & \includegraphics[scale=0.5]{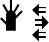} & \includegraphics[scale=0.5]{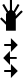} & \includegraphics[scale=0.5]{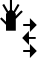} & \includegraphics[scale=0.5]{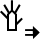} \\
    \end{tabular}
    }
    \caption{Top 10 nearest neighbors for selected signs using different evaluation metrics. The reference signs are shown at the top, and the retrieved signs are displayed in order of decreasing similarity score from top to bottom.}
    \label{tab:nearest_neighbors}
\end{table*}

\begin{itemize}
    \item \textbf{\texttt{BLEU}} often misses the closest signs, due to its symbol order sensitivity and exact token match requirements.
    \item \textbf{\texttt{chrF}} performs slightly worse than modified \texttt{BLEU}, focusing on character-level overlaps but missing key patterns and visual similarities.
    \item \textbf{\texttt{CLIPScore}} often fails to differentiate subtle yet fundamental changes and assess closeness, likely due to the model's domain mismatch and tendency to produce similar embeddings.
    \item \textbf{Symbol Distance} consistently retrieves signs that are \emph{more} visually and semantically similar to the reference, showing its effectiveness in capturing SignWriting's nuances.
\end{itemize}

These qualitative results highlight the limitations of traditional metrics such as \texttt{BLEU}, \texttt{chrF} and \texttt{CLIPScore} for SignWriting, and demonstrate the advantage of utilizing a metric specifically designed to account for the properties of SignWriting.




\section{Conclusions}

In this paper, we introduced a comprehensive toolkit, \texttt{signwriting-evaluation}, providing specialized evaluation metrics for SignWriting strings. Our proposed Symbol Distance Metric is specifically designed to capture the unique characteristics of SignWriting, addressing the limitations of traditional metrics like \texttt{BLEU}, \texttt{chrF}, and \texttt{CLIPScore} when applied to sign language data.

Through qualitative evaluations, we demonstrated that our Symbol Distance Metric outperforms other metrics when assessing the similarity of SignWriting signs. It effectively accounts for symbol attributes, positions, and variations in symbol ordering, providing a more accurate and meaningful evaluation of model outputs.

We, therefore, encourage the research community to adopt our Symbol Distance Metric for evaluating SignWriting models, as it offers significant improvements over standard metrics not tailored for sign language processing. Using more appropriate tools will allow researchers to develop and assess models more effectively, ultimately advancing the field of sign language technology.

Our open-sourced toolkit is available at \url{https://github.com/sign-language-processing/signwriting-evaluation}. We welcome community contributions to enhance the toolkit, develop new metrics, and collaborate on advancing SignWriting evaluation, creating useful tools benefiting both researchers and the deaf community.

\section*{Limitations}
We recognize that our current evaluation primarily emphasizes qualitative analysis. To further validate our metrics, a more rigorous quantitative analysis, aligned with human judgments, is necessary. This would provide deeper insights into how accurately the metric captures sign similarity and meaning.

There is also room for future enhancement. Incorporating machine learning techniques could optimize the weighting of symbols and better capture the subtle complexities of SignWriting. 

Additionally, although our toolkit supports continuous signing by treating the sign sequence as a set and identifying the minimal match, expanding these metrics to better handle multi-sign sequences remains an important area for future research.

\bibliography{background,anthology,custom}
\bibliographystyle{acl_natbib}

\end{document}